\documentclass[letterpaper, 10 pt, journal, twoside]{ieeetran}

\IEEEoverridecommandlockouts                              

\usepackage{graphicx}
\usepackage{balance}

\newcommand{\CHANGED}[1]{{\color{black} #1}}
\usepackage{pifont}

\let\labelindent\relax

\usepackage{amsmath,amssymb,amsthm}
\usepackage{graphicx}
\usepackage{xspace}
\RequirePackage[dvips]{epsfig} 
\RequirePackage{float}
\RequirePackage{fancyhdr}
\RequirePackage{pstricks,pst-plot,psfrag}
\RequirePackage{multirow}
\RequirePackage{rotating}
\RequirePackage{units}
\usepackage{booktabs}
\usepackage{pgfplots}
\newlength\figureheight
\newlength\figurewidth
\pgfplotsset{compat=1.14}
\usepackage{array}
\usepackage{subcaption}
\usepackage{setspace}
\usepackage{enumitem}
\usepackage{hyperref}

\newcommand{\newsec}[1]{\vspace{2mm} \noindent \textbf{#1.} }

\title{
Zoomorphic Gestures for Communicating\\ Cobot States
}

\author{Vanessa Sauer$^{1}$, Axel Sauer$^{2}$, and Alexander Mertens$^1$

\thanks{Manuscript received: October, 14, 2020; Revised: January, 2, 2021; Accepted: February, 8, 2021.} 
\thanks{
This paper was recommended for publication by Editor Gentiane Venture upon evaluation of the Associate Editor and Reviewers' comments.} 
\thanks{$^{1}$ Vanessa Sauer and Alexander Mertens are with the Institute of Industrial Engineering and Ergonomics at RWTH Aachen University, Germany.
        {\tt\small v.sauer@iaw.rwth-aachen.de, a.mertens@iaw.rwth-aachen.de}}%
\thanks{$^{2}$ Axel Sauer is with the Autonomous Vision Group at the Max-Planck Institute for Intelligent Systems and the University of Tübingen, Germany. This work was done while Axel Sauer was with the Technical University of Munich.
        {\tt\small axel.sauer@tue.mpg.de}}%

\thanks{Digital Object Identifier (DOI): see top of this page.}
}

\begin{document}

\maketitle

\begin{abstract}
Communicating the robot state is vital to creating an efficient and trustworthy collaboration between humans and collaborative robots (cobots).
Standard approaches for Robot-to-human communication face difficulties in industry settings, \CHANGED{e.g., because of high noise levels or certain visibility requirements}.
Therefore, this paper presents zoomorphic gestures based on dog body language as a possible alternative for communicating the state of appearance-constrained cobots. 
For this purpose, we conduct a visual communication benchmark comparing zoomorphic gestures, abstract gestures, and light displays.
We investigate the modalities regarding intuitive understanding, user experience, and user preference. 
In a first user study $(n = 93)$, we evaluate our proposed design guidelines for all visual modalities.
A second user study $(n = 214)$ constituting the benchmark indicates that intuitive understanding and user experience are highest for both gesture-based modalities. 
Furthermore, zoomorphic gestures are considerably preferred over other modalities.
These findings indicate that zoomorphic gestures with their playful nature are especially suitable for novel users and may decrease initial inhibitions.
\end{abstract}


\section{Introduction}
\IEEEPARstart{I}{nteraction}
and, more specifically, communication between humans and robots are central challenges in collaborative robotics \cite{Barattini.2012}. While human-to-robot communication is an intensively researched field, the inverse direction, communication from robot to human (RtH), has received less attention in comparison \cite{fulton2019robot}. 
Yet, the communication of the robot's state and the acknowledgment of user commands are required for effective and safe collaboration \cite{BarakaVeloso.2018}.
Well-designed communication can also increase trust levels and lead to a more engaging interaction with the collaborative robot (cobot) \cite{BarakaVeloso.2018, Aubert.2018}. 
The system status has to be informative and easy to understand to meet those demands.

According to Onnasch et al. \cite{Onnasch.2016}, cobots can communicate with human users via acoustic, mechanical, or visual modalities. In an industry setting, acoustic communication may be impeded due to the level and spectral characteristics of noise prevalent in these settings. Alternatively, acoustic RtH communication would have to be designed very intrusively, e.g., via loud warning tones \cite{BarakaVeloso.2018}. Mechanical RtH communication may limit the range in which a user can perceive feedback, such as vibrations. Haptic interfaces can expand this space but require additional hardware and the willingness of human users to wear the interface while interacting with the cobot. Several modalities are available in visual communication, such as text displays, lights, or gestures. Compared to the other visual modalities, gestures have the advantage of being visible from many different positions and distances and do not require any additional hardware \cite{BarakaVeloso.2018}.

For these reasons, we specifically focus on gestures for RtH communication in an industry setting.
Gestures for cobots are commonly based on human body language \cite{fulton2019robot} due to an easier understanding by the user.
\CHANGED{In this work, we focus on} functionally designed,
appearance-constrained cobots \CHANGED{lacking expressive faces} \cite{Bethel.2007}, i.e., robot arms 
without specific humanoid design characteristics.
\CHANGED{For these types of cobots, we argue that} gestures based on human body language may raise excessive expectations regarding the cobot's capabilities \cite{Syrdal.2010}. 
Furthermore, human-inspired gestures for a non-humanoid cobot may create uneasiness due to a perceptual mismatch of sensory cues (cobot appearance vs. gesture design) \cite{Kaetsyri.2015}. 
Instead, \textit{zoomorphic gestures} based on animals' body language may be an alternative approach for gesture design.

Our proposed approach of zoomorphic gestures for industry cobots may offer the advantages of gesture communication (intuitive, visibility from different positions, no additional hardware) while avoiding exaggerated expectations and perceptual mismatch. In this work, we investigate if zoomorphic gestures provide an intuitive understanding, i.e., whether the modality's meaning can be unambiguously understood \cite{deshmukh2018more}.
Furthermore, we evaluate the user experience (attractiveness, joy of use, intuitive use, and intent to use) and the user preferences compared to other visual communication modalities, such as light displays and abstract gestures.

In this work, we offer several contributions to the research domain of RtH communication:
\begin{itemize}
    \item With zoomorphic gestures, we offer a novel approach of RtH communication for appearance-constrained cobots.
    \item We propose design guidelines for the development of zoomorphic gestures and \CHANGED{evaluate} them in an \CHANGED{online} user study ($n=93$).
    \item We conduct a benchmark of three visual modalities for communicating robot states.
    We perform the benchmark with a large-scale online user study ($n=214$) by recruiting participants with diverse demographics.
\end{itemize}

\section{Related Work}
RtH communication in industry settings is required to help human users predict the cobot's behavior and actions more accurately \cite{Aubert.2018}. Communication of the robot state is possible via different channels \cite{Onnasch.2016}. Due to the industry context, we focus on non-facial, non-humanoid communication of the robot state and feedback to the user while omitting the display of robot emotions. 
Based on the limitations of acoustic (e.g., high noise levels in industry settings) and haptic (e.g., required hardware, user acceptance) RtH communication in industry settings, visual communication modalities are preferred for industry cobots.

\newsec{Visual RtH communication}
Light displays have a long history of being used to express the state of an electronic device in general or of a robot state in particular. Yet, the design space and the light codes 
are often kept simple.
This simplification reduces the learning effort by utilizing
typical signals of widespread electronic devices \cite{Harrison.2012, Cha.2017}. Nonetheless, lights to communicate the robot intent and state have been explored in industry settings \cite{Chadalavada.2015, Unhelkar.2014} and other applications such as mobile or flying robots \cite{Baraka.2016, Szafir.2015}. 
Light displays have the advantage of being visible from different distances and angles \CHANGED{without interfering with the task execution}. They also provide persistent information compared to the transient nature of acoustic feedback \cite{BarakaVeloso.2018}. However, the understandability and usefulness of light displays largely depend on the specific design and placement on the robot, requiring time and effort in the design process \cite{Cha.2017}.

Previous work on RtH communication in industry settings also includes approaches where the robot response and state is displayed on a screen. 
The robot state can be visualized using pictograms \cite{Aubert.2018} or text-based information \cite{ElMakrini.2017}. Moreover, communication of robot faults with augmented reality has also been explored \cite{DePace.2018}.
While explicit screen-based communication tends to be straightforward and easy to interpret, user studies investigating the use of screens show that this communication modality may lead to information overload \cite{Aubert.2018}. \CHANGED{Moreover, this modality may be less visible from a distance, especially for small screens or texts.}

Visual RtH communication can also utilize motion-based communication, which primarily entails communication via gestures \cite{Aubert.2018}. Human-inspired gestures are dominant in industry and related applications as they are easy to understand \cite{fulton2019robot}, especially when viewed within the task context \cite{gleeson2013gestures}. Human-inspired gestures have been used for assembly tasks \cite{ ElMakrini.2017, gleeson2013gestures} but also less related fields such as underwater robots \cite{fulton2019robot}.
Further, a combination of human-like movements with a humanoid design of the physical robot can reduce users' stress \cite{Zanchettin.2013}.
However, appearance-constrained cobots do not necessarily have the required physical design to rely on human cues or gestures to communicate the robot states like humanoid cobots \cite{Cha.2017}. Moreover, perceptual mismatch \cite{Kaetsyri.2015} and exaggerated expectations of the cobot's capabilities \cite{Syrdal.2010} can arise with human-inspired gestures. 

Alternatively, gestures for motion-based communication may also be abstract.
In this approach, \CHANGED{novel movements, i.e., not based on any body language,} are \CHANGED{developed} and connected with semantics \CHANGED{to express} certain robot states or feedback. 
\CHANGED{In the following, we refer to those movements as \textit{abstract gestures}.
We assume that the trajectory of these gestures is optimal with respect to a particular efficiency metric, creating a robot-typical impression. 
This metric may be any of the ones commonly used in trajectory planning literature, e.g., minimum execution time, minimum energy, or minimum jerk \cite{Gasparetto2012Trajectory}.}
\cite{Aubert.2018} and \cite{Dragan.2015} explored the use of robot motion to express robot intent abstractly. Instead of developing dedicated gestures, they investigate how they can alter goal-oriented movements to include legible information on the robot's objective. 
\CHANGED{In a similar vein, 
\cite{knight2016expressivediss}
utilize methods from drama and dance, i.e., Laban movement theory, to derive expressive motions for appearance-constrained, mobile robots. 
\cite{takahashi2010remarks} apply the same approach to a social robot to communicate affect.
\cite{fernandez2018passive} use passive demonstrations to communicate navigational intentions of a mobile robot.
Moreover, \cite{Venture2019robot} provide general recommendations for realizing expressive motions based on the robot's morphology and the desired movement.}
Abstract gestures have the advantage that they can be tailored to the physical capabilities of an appearance-constrained cobot. 
However, they \CHANGED{may} come at the expense of increased effort to learn the new gestures and their respective semantics, which can increase the cognitive load \cite{Aubert.2018}.

\newsec{Zoomorphic gestures}
To address the issues of the communication modalities mentioned above, zoomorphic gestures, based on animal body language, pose one possible approach.
To our knowledge, zoomorphic gestures have not yet been considered for RtH communication of cobots 
\CHANGED{for industrial applications such as manufacturing.}
However, they have found applications in social robotics, including appearance-constrained robots. In this field of application, the focus lies on affective displays instead of solely communicating information on the functional state \CHANGED{-- for} example, adding a dog tail to a utility robot \cite{Singh.2013}, or leveraging dog-based gestures \cite{Syrdal.2010, Gacsi.2016}.
Dog-based gestures can achieve high classification accuracy, especially when viewed within the task context \cite{Gacsi.2016}.
Zoomorphic interaction is not limited to RtH communication but also extends to human-to-robot communication, e.g., by using a dog-leash interface to direct a robot \cite{Young.2011}.
The dominant use of dog body language for zoomorphic gestures can be traced back to the close relationship between dog and human, which leads to an intuitive understanding of dog-gestures even by people who do not own a dog \cite{Syrdal.2010}.

%
%
\section{Methods}
We conducted a benchmark of zoomorphic gestures, abstract gestures, and light displays. We use the benchmark to explore if zoomorphic gestures provide a better intuitive understanding and user experience for appearance-constrained cobots than other visual modalities. We use the following approaches and methods to implement the three visual communication modalities and conduct the benchmark. 

\begin{figure}
    \centering
    \includegraphics[width=\linewidth]{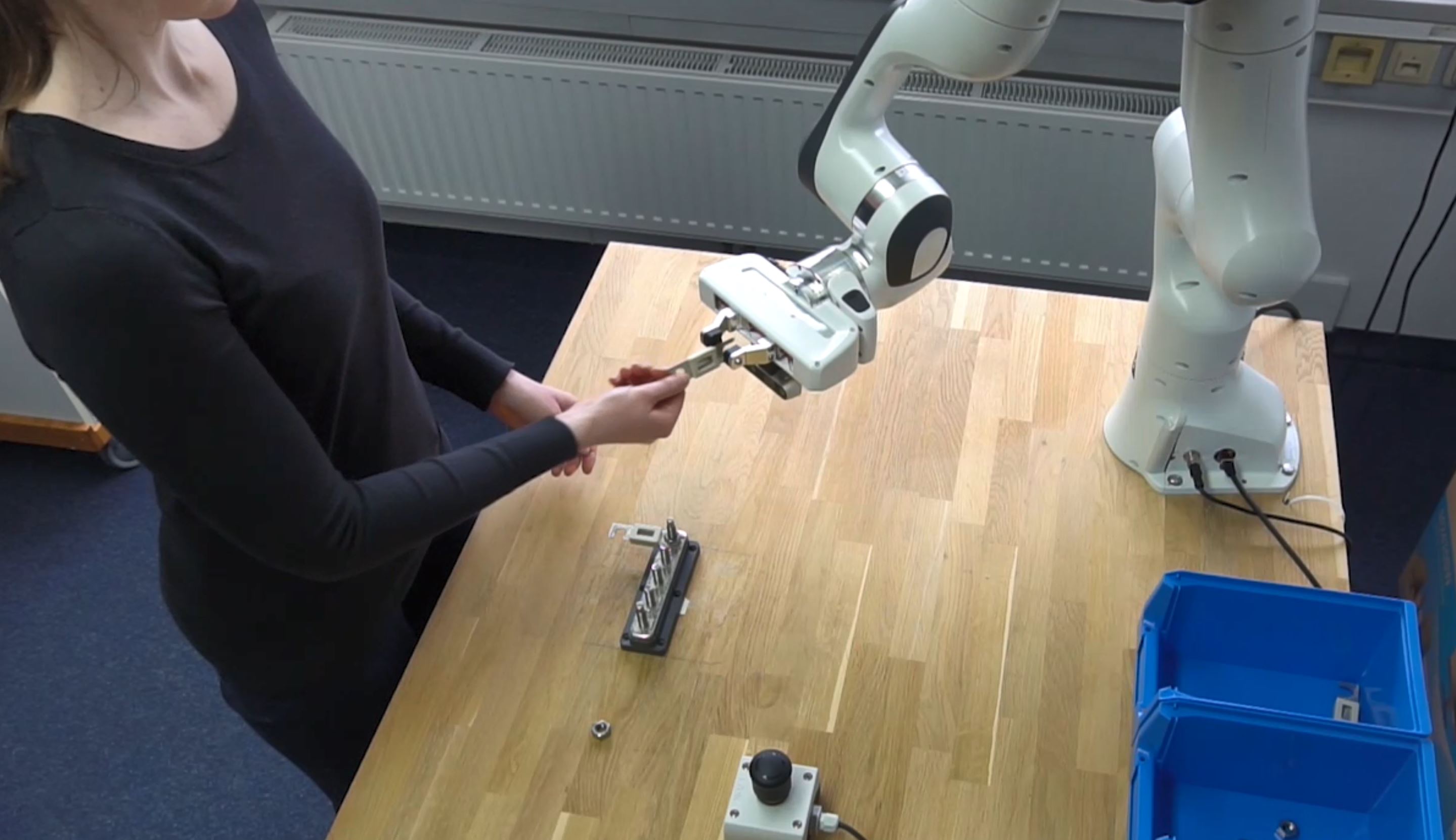}
    \caption{\textbf{Collaborative Assembly Task.}
    We focus on the collaborative assembly of fuses as the scenario for gesture development and user evaluation.
    }
    \label{fig:scenario}
\end{figure}

\newsec{Robot system}
We use the collaborative robot system Panda by Franka Emika (see Fig. \ref{fig:scenario}) to implement the gestures and light displays and to perform the subsequent user studies. Panda is a functional robot targeted at lightweight collaborative assembly tasks. 

\newsec{Application task and use cases}
For a concrete application focus, we defined a collaborative task from which we derived five use cases. As cobots are likely to be used in collaborative assembly, we chose an assembly task in which the cobot is responsible for the part acquisition. Concretely, the cobot acquires fuses from a storage container and delivers them to the users (see Fig. \ref{fig:scenario}). 
The human performs part manipulation (inserting the fuse into a fuse holder) and part operation (screwing the nut to secure the fuse) \cite{gleeson2013gestures}. 

Based on this specific assembly task and related studies on similar tasks, e.g., \cite{ElMakrini.2017}, we derived the following five use cases:
(i) Greet user,
(ii) Prompt user to take the part,
(iii) Wait for a new command,
(iv) Error: Storage container is empty,
(v) Shutdown.

\begin{figure}
    \centering
    \includegraphics[width=1\linewidth]{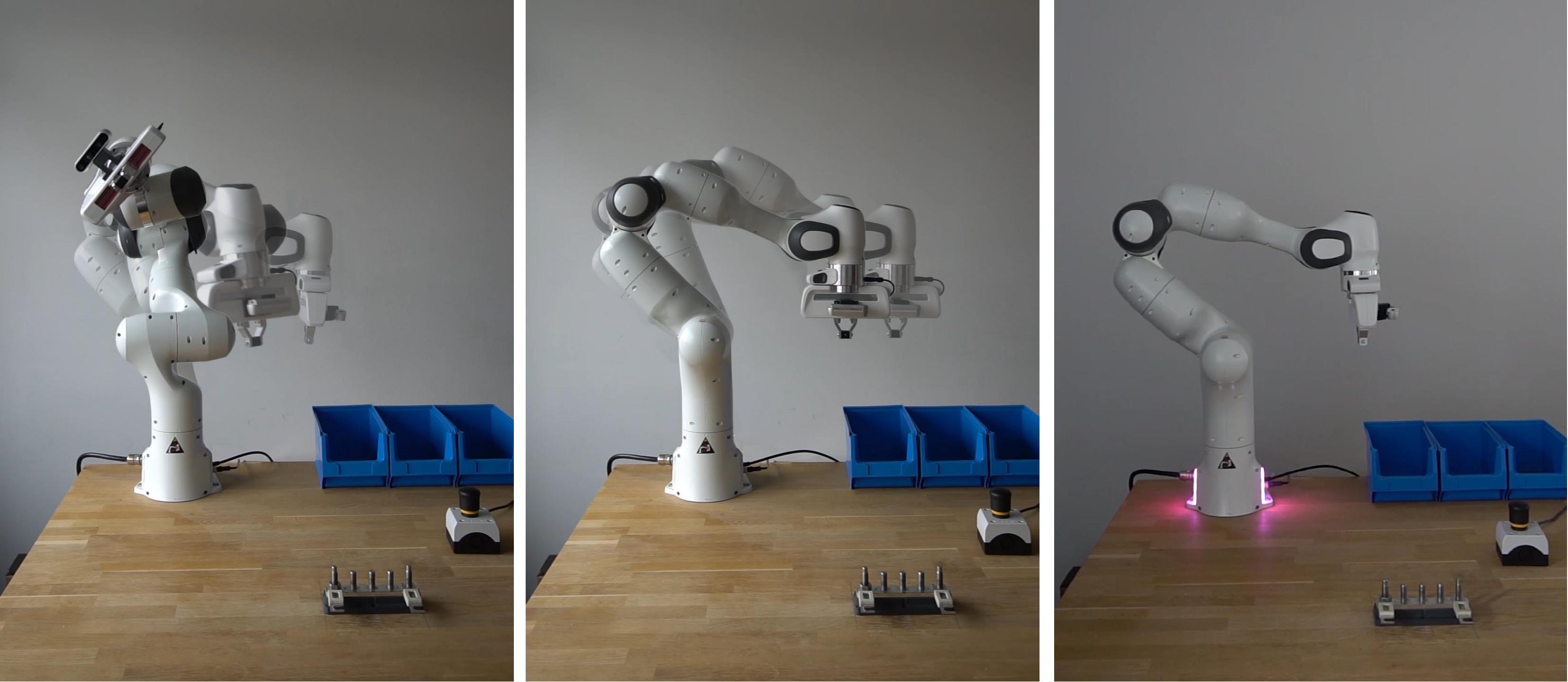}
    \caption{\textbf{Benchmark Modalities.}
    We show the modalities evaluated in the communication benchmark for the use case "error: storage container empty". From left to right: zoomorphic gesture, abstract gesture, light display. \CHANGED{We illustrate the movements with faded previous positions.}
    }
    \label{fig:modalities}
\end{figure}

\newsec{Modalities}
We compare three different modalities (see Fig. \ref{fig:modalities}). Both the zoomorphic and abstract gestures aim to develop emblems \cite{Knapp.1972} based either on dog-body language (zoomorphic gesture) or by developing new, self-contained movements that convey a specific meaning or robot state (abstract gestures).

\textit{Zoomorphic gestures.}
For each use case, \CHANGED{we mirrored the intention of the robot (e.g., prompting the user to take a part) to an intention, a dog may have (e.g., encouraging the owner to play). In a second step, we collected gestures that dogs use to express the respective intention by leveraging real-life interaction with dogs, online videos, and literature \cite{Singh.2013}.}
We then translated the dog gestures into three distinct zoomorphic gestures
\CHANGED{by jointly applying }the following guidelines inspired by \cite{fulton2019robot}:
\begin{itemize}
    \item \textit{Mimicry.} We mimic specific dog behavior and body language to communicate robot states.
    \item \textit{Exploiting structural similarities.} 
    Although the cobot is functionally designed, we exploit certain components to make the gestures more "dog-like," e.g., the camera corresponds to the dog's eyes, or the end-effector corresponds to the dog's snout
    \item \textit{Natural flow.} We use kinesthetic teaching and record a full trajectory to allow natural and flowing movements with increased animacy.
\end{itemize}

\textit{Abstract gestures.}
To omit the semantics of human or animal body language in designing the abstract gestures, we created new, cobot-specific gestures inspired by stereotypical robots and consumer electronics. 
Towards this goal, we adopted the following guidelines:
\begin{itemize}
    \item \textit{Simplicity.} The gestures should generally be simple and goal-oriented.
    \item \textit{Efficiency.} 
    \CHANGED{    
    Again, we leverage kinesthetic teaching but only record several keyframes \cite{akgun2012trajectories}. We utilize a minimum-jerk trajectory planner to generate efficient trajectories between the keyframes. At each keyframe, the robot remains for a moment before moving on, leading to a less flowing and more robot-typical movement.}
\end{itemize}
As before, we create three different abstract gesture options per use case.

\textit{Light display.}
The light display serves as a \CHANGED{reference modality} for the two gesture-based modalities in the benchmark. \CHANGED{We chose light displays as a reference, because they are a widespread and accepted form to communicate the state of electronic devices in general and robots in particular \cite{Harrison.2012}.}
We applied the \CHANGED{guidelines} by \cite{fulton2019robot} to the design space of Franka Emika Panda's built-in light display to make the light display as robust and intuitive as possible:
\begin{itemize}
    \item \textit{Light color.} Color corresponds to the general status: white (neutral), green (prompt for interaction), red (error), light off (robot off). \CHANGED{We selected additional light colors (white and light off) compared to \cite{fulton2019robot}.}
    \item \textit{Blinking frequency.} Higher frequency indicates a more urgently required reaction by the user.
    \item \textit{Similarity.} Similar robot states are communicated with similar light displays, \CHANGED{i.e., color and blinking frequency.}
\end{itemize}
We recorded the different states of the cobot's built-in lights.
We combined the recordings using video editing software to create the desired light codes resulting in three options per use case.

\newsec{Experiments}
We conduct two user studies to reduce degrees of freedom in the design process and to examine whether zoomorphic gestures are more intuitive and attractive than other visual modalities. The first user study aims at identifying the best option to communicate the robot state out of the three designs per use case and modality. Further, we also evaluate the suitability of the design guidelines outlined above. The second user study utilizes the best option per use case and modality identified in the first study to assess the three modalities regarding intuitive understanding, user experience, and stated preference.
 
\section{Study I: Identifying Best Design Options}
To identify which of the three design options per use case is the best way to communicate the robot state with a given modality, we utilized an online user study. As the study design does not require direct interaction with the cobot, an online study is a considerably more efficient tool to collect as many responses as possible compared to in-person studies.

\newsec{Population}
We recruited a convenience sample of 93 participants (39 male, 52 female, 1 other, 1 N/A, average age $M=27.84$ years, $SD=8.33$). The participants mainly had a non-technical background (63 non-technical, 26 technical, 4 other) and, on average, a positive attitude towards robots in general ($M=3.91, SD=0.89$, rated on a scale from 1 (very negative) to 5 (very positive)).

\begin{figure}
    \centering
    \includegraphics[width=1\linewidth]{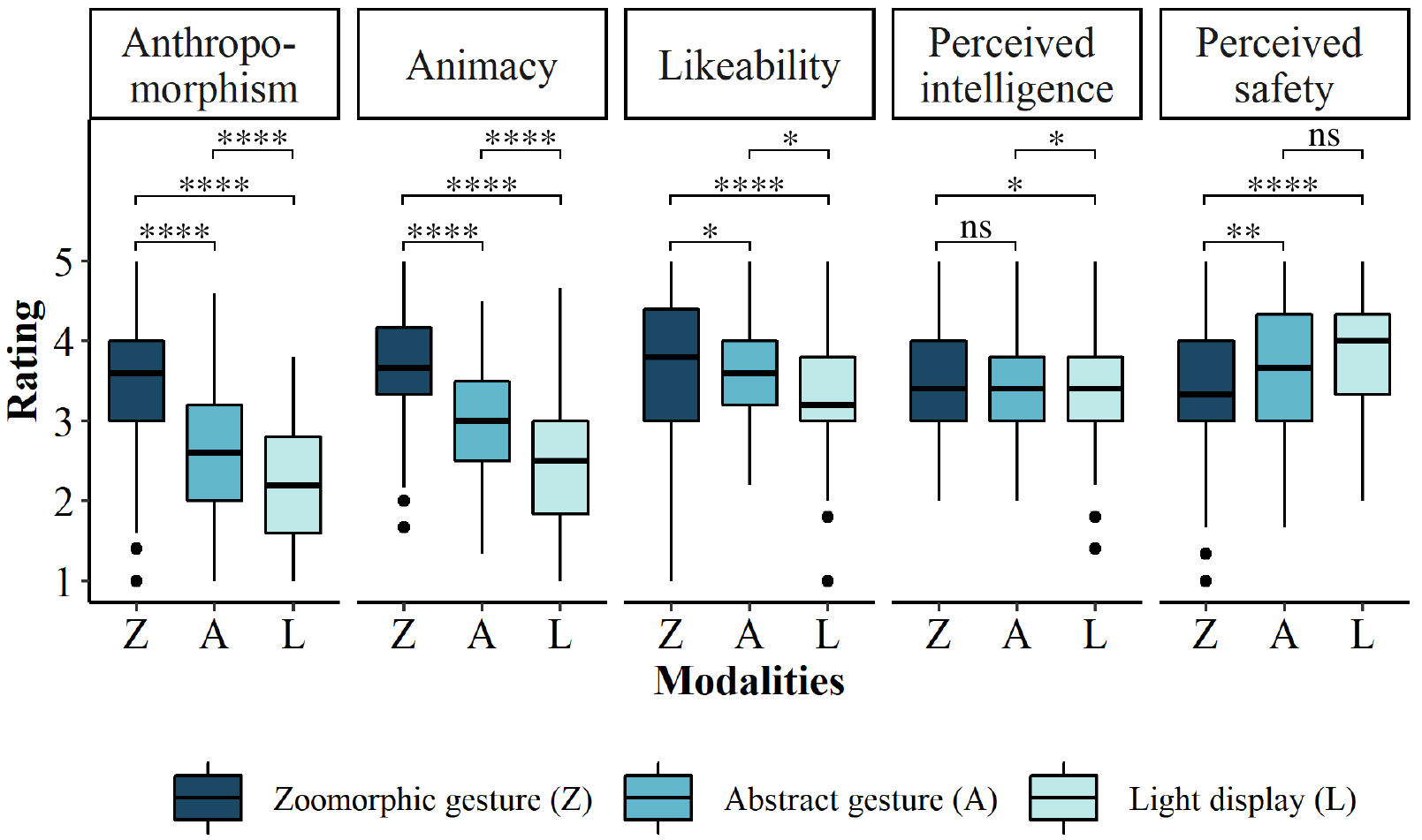}
    \caption{\textbf{Godspeed Ratings.}
    Boxplot of the ratings of the three modalities over the five dimensions of the Godspeed questionnaire.
    We also report the Bonferroni-corrected pairwise comparisons:
    ns: $p>0.05$, 
    *: $p<0.05$, 
    **: $p<0.01$, 
    ***: $p<0.001$, 
    ****: $p<0.0001$, 
    }
    \label{fig:ANOVA1}
\end{figure}

\newsec{Experimental procedure}
After giving informed consent, we provided information on the assembly task (description and video, \CHANGED{see Fig. \ref{fig:scenario} for a screenshot}) to the participants. For each modality and use case, the participants viewed three videos showing the three design options. We then asked the participants which option fits best to communicate the given robot state. 
After completing all five use cases within one modality in a randomized order, the participants evaluated the robot communicating with the given modality.
We used the Godspeed questionnaire \cite{bartneck2009measurement} and asked for free-text associations the participants had with the shown modality.
The Godspeed questionnaire was developed
as a standardized measure for the impression a robot has on the user
and considers five dimensions (see Fig. \ref{fig:ANOVA1}) \cite{bartneck2009measurement}. Although other measures exist, the Godspeed questionnaire is widely utilized and provides sufficient statistical reliability and validity \cite{bartneck2009measurement, Weiss.2015}.
The blocks of each modality were randomized as well to avoid anchoring and sequence effects \cite{Fanning.2005}. The online study concluded with a short demographic questionnaire. 

\begin{figure}
    \centering
    \includegraphics[width=1\linewidth]{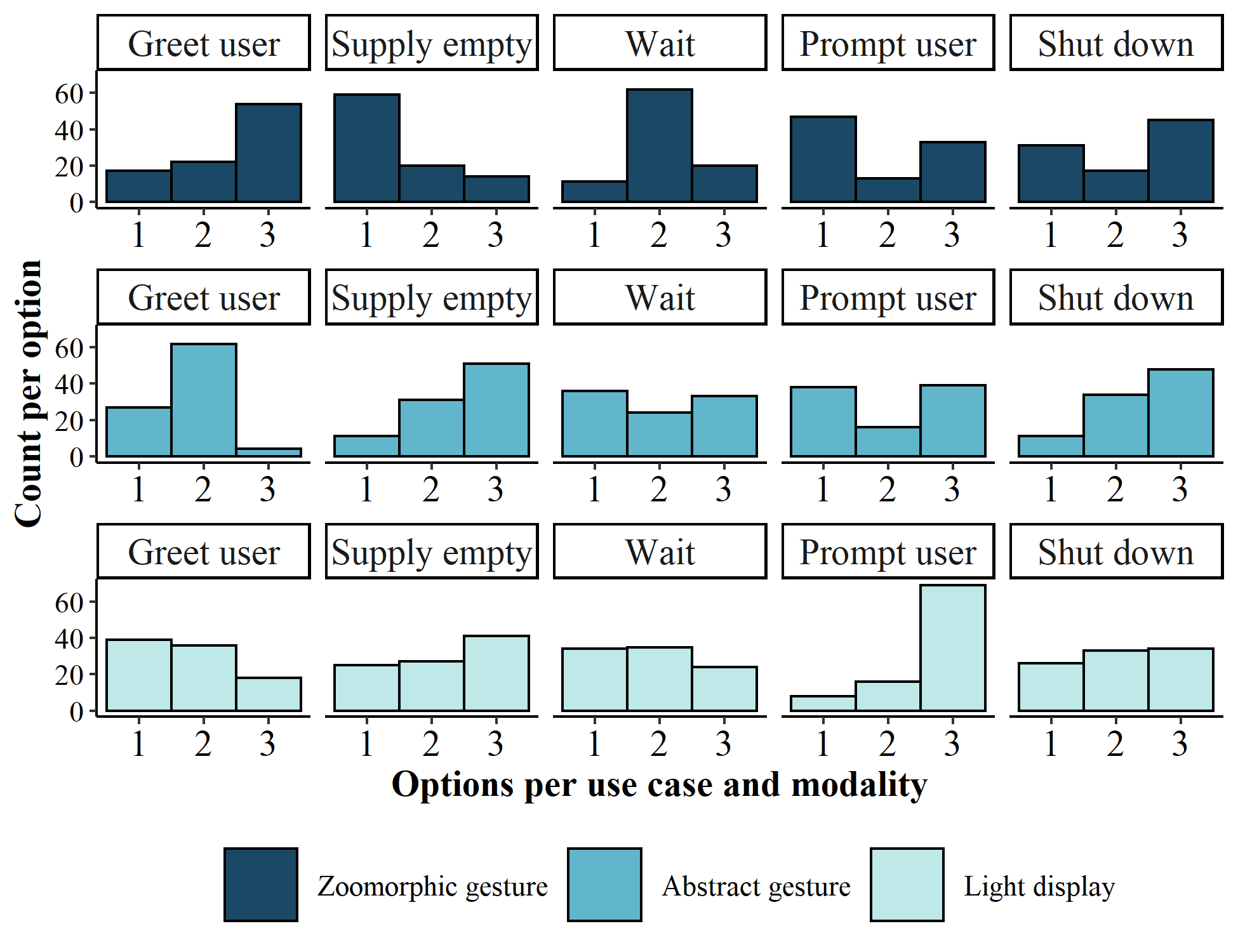}
    \caption{\textbf{Choice Distributions.}
    For each use case and modality we designed and evaluated three distinct options.
    }
    \label{fig:Histo1}
\end{figure}

\newsec{Results}
The results in Fig. \ref{fig:Histo1} indicate that for zoomorphic gestures, a clear best option generally exists.
The ambigousness is higher in the case of abstract gestures and light displays. For these modalities, several use cases have two or even all three design options selected similarly often, e.g, use case "wait".

We analyze the evaluation of the modalities with the Godspeed questionnaire with one-way analyses of variance (ANOVA), see Fig. \ref{fig:ANOVA1}. We report the F-statistic ($F$), the p-value ($p$) and the effect size ($\eta_{G}^2$). The results indicate that zoomorphic gestures score significantly higher than the other modalities in the dimensions 
anthropomorphism 
$(F(2,184)=95.57$, $p < 0.001$, $\eta_{G}^2 =0.305)$, 
animacy 
$(F(2,184)=95.07$, $p < 0.001$, $\eta_{G}^2 =0.309)$, 
and likeability 
$(F(2,184)=13.01$, $p < 0.001$, $\eta_{G}^2 =0.062)$.
In the case of perceived intelligence, both gesture-based modalities are rated significantly higher than the light displays 
$(F(2,184)=5.62$, $p= 0.004$, $\eta_{G}^2 =0.020)$. 
Zoomorphic gestures are rated significantly lower in regards to perceived safety than the alternatives 
$(F(2,184)=12.88$, $p < 0.001$, $\eta_{G}^2=0.046)$.

The participants' free-text answers reveal strong associations of the zoomorphic gestures with dogs (named by 28 participants) and with other animals (20) or humans (15).
For the abstract gestures, strong associations with robots (21) or machines (10)  but only rarely with dogs (5) were expressed. Cobots communicating via light displays were associated with traffic/control lights (36) or electronics in general (20), but hardly with dogs (0) or any other living beings.

\newsec{Discussion}
The different levels of embedded semantics may explain the varying degree of ambiguousness over the best design option for a particular modality. In the case of zoomorphic gestures, semantics are transferred from dog body-language, which tends to be easy to understand both by dog-owners and people with little to no experience with dogs \cite{Syrdal.2010}. 
Given that abstract gestures and light displays do not have an apparent semantic connection, participants may have chosen the best option per use case and modality more along personal preference ("What looks nice or appealing?").

The \CHANGED{Godspeed ratings,} along with the voiced associations, \CHANGED{confirm} the functionality of our proposed design guidelines. \CHANGED{For example, zoomorphic gestures received high ratings for anthropomorphism and animacy along with frequent dog associations. Abstract gestures were rated less anthropomorphic and animate, and were frequently associated with robots and machines.}
\CHANGED{Thus, }the guidelines are suitable for integrating typical dog-characteristics into the zoomorphic gestures and help create robot-typical abstract gestures. The participants' comments indicate that the light blinking frequency was not always understood as an indication of urgency. Instead, some participants interpreted blinking as a processing state of the cobot during which the user needed to wait. 
Our recommendation for future studies is to review the use of blinking and its frequency.
The modalities were rated similarly regarding perceived intelligence. 
This finding may be explained by the fact that the cobot's general reaction was independent of the modality. For example, the cobot always reported an error over an empty storage bin.
In terms of perceived safety, zoomorphic gestures are perceived as less safe than the alternatives, although the effect size is small $(\eta_G^2=0.046)$. 
Nonetheless, it is not surprising that zoomorphic gestures cause more anxious, agitated, and surprised reactions than the alternatives, as zoomorphic gestures are less robot-typical and more extensive. When viewed globally, zoomorphic gestures are still rated above the scale middle ($M=3.41, SD=1.24$, see Fig. \ref{fig:ANOVA1}), hence, achieve sufficient subjective safety ratings.

Overall, the results of the Godspeed questionnaire indicate that the modality used to communicate the robot state shapes the user's perception of the robot but to different extents. The modality has substantial effects on anthropomorphism and animacy. Likability, perceived intelligence, and perceived safety are less affected.
Additionally, this study identifies the best option of the design alternatives per use case and modality, a prerequisite for the following benchmark study.

\section{Study II: Visual Communication Benchmark}
After identifying the best design option per use case for each modality, we performed a second user study -- the visual communication benchmark.

\newsec{Population}
In total, 214 participants (116 male, 96 female, 2 other, average age $M = 40.44$ years, $SD=12.69$) were recruited for the online study via an online panel. The participants mainly had a non-technical background (110 non-technical, 79 technical, 25 other) and, on average, a positive attitude towards robots in general ($M=4.04, SD=0.93$, rated on a scale from 1 (very negative) to 5 (very positive)). The sample included participants with low prior experience (81 participants with no practical experience) and high prior experience (60 participants with practical experience with at least three robot types) with robots from five different domains (industry, household, entertainment, service, healthcare).

\newsec{Experimental procedure}
After giving informed consent and providing relevant demographic information, information on the assembly task was provided in the same way as in the first study (description and video). For each modality and use case, the participants viewed the video showing the best option identified in study I. The participants were asked which option from a predefined list describes best what the cobot wanted to express in each video. The predefined list remained the same for each use case and all modalities. It comprised the five use cases plus two additional ones (cobot wants to check assembled product, cobot is malfunctioning). The additional use cases prohibit an obvious mapping. 
After completing all five use cases within one modality in a randomized order, the participants were asked to rate the robot communicating with the given modality regarding user experience (intuitive use, joy of use, attractiveness \cite{Schrepp}) and intent to use \cite{Harborth.2018}. The blocks of each modality were randomized as well to avoid anchoring and sequence effects \cite{Fanning.2005}. The online study concluded by asking the participants to state their preference for one of the three modalities.

\begin{figure}
    \centering
    \includegraphics[width=1\linewidth]{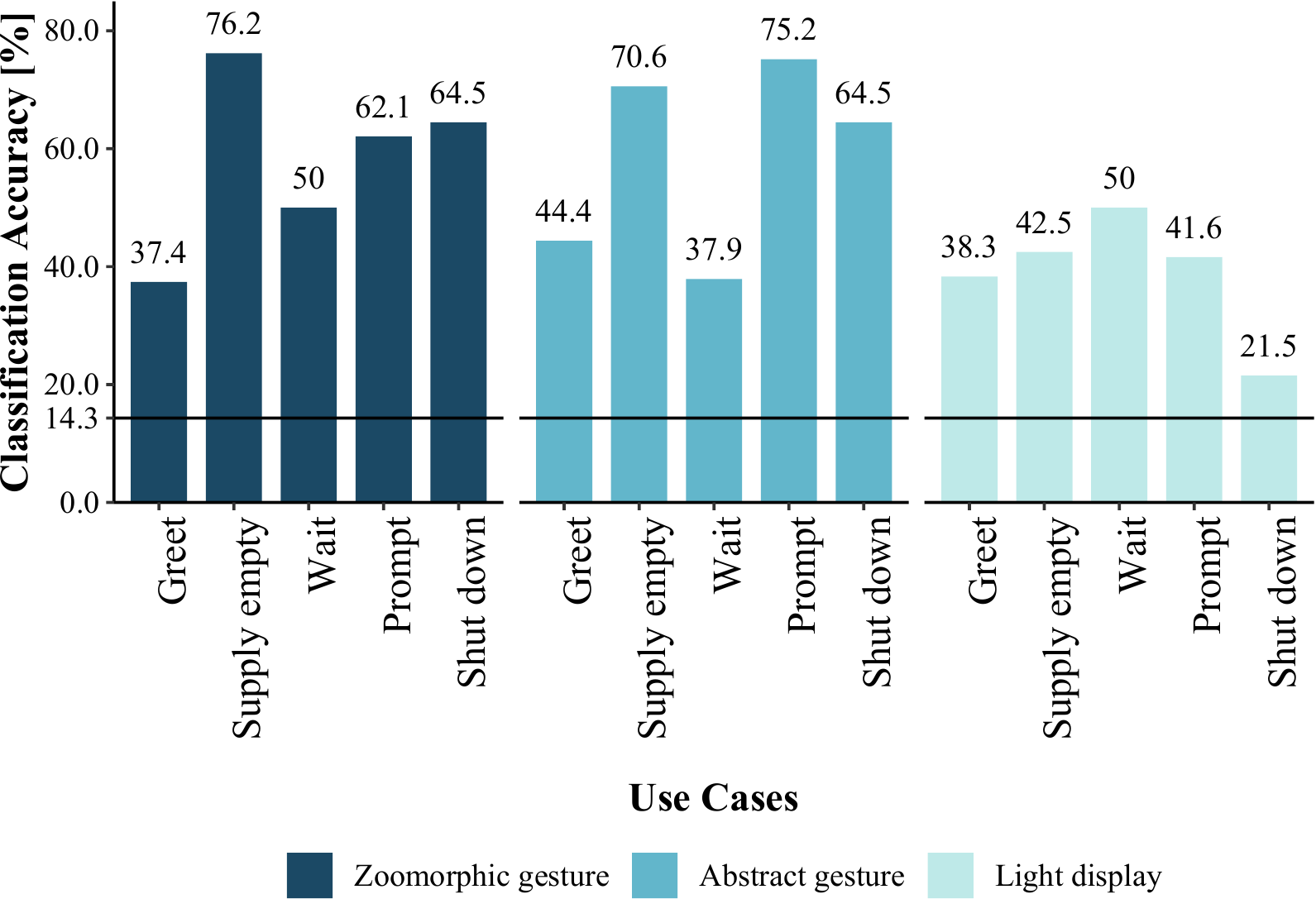}
    \caption{\textbf{Classification Accuracy per Modality and Use Case.} The classification accuracy is the number of correct answers divided by the total number of participants.
    The chance level is at $14.3 \%$, as indicated by the black lines.
    }
    \label{fig:Histo2}
\end{figure}

\begin{figure}
    \centering
    \includegraphics[width=1\linewidth]{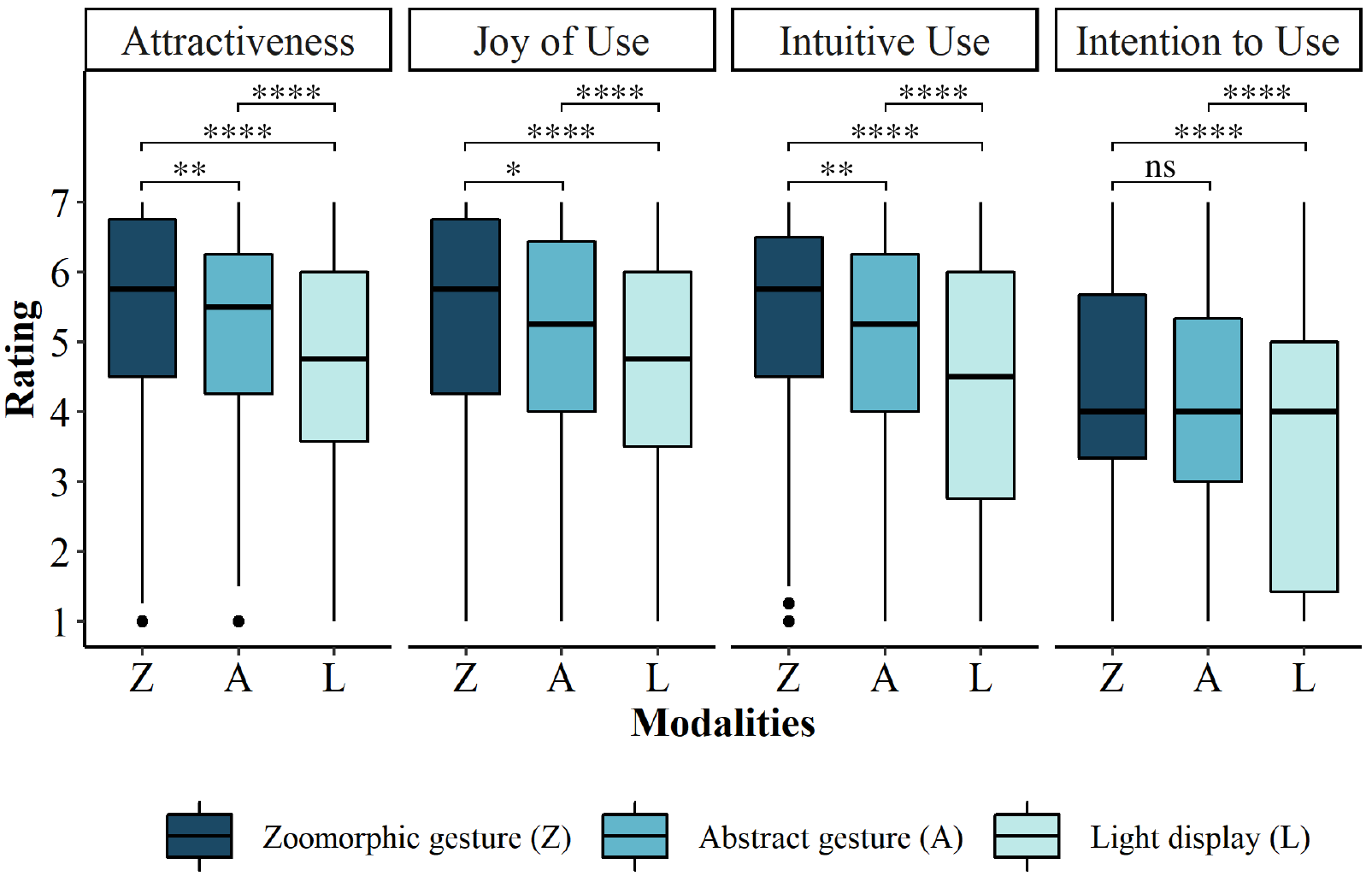}
    \caption{\textbf{User Experience Ratings.}
    Boxplot of the ratings of the three modalities over the four user experience dimensions.
    We also report the Bonferroni-corrected pairwise comparisons:
    ns: $p>0.05$, 
    *: $p<0.05$, 
    **: $p<0.01$, 
    ***: $p<0.001$, 
    ****: $p<0.0001$, 
    }
    \label{fig:ANOVA2}
\end{figure}

\newsec{Results}
To evaluate the understandability of the different modalities we consult the classification accuracy \cite{deshmukh2018more} (see Fig. \ref{fig:Histo2}). 
On average, the classification accuracy reaches 58.0\% for zoomorphic gestures, 58.5\% for abstract gestures, and 38.8\% for light displays. 
A one-way ANOVA with Bonferroni-adjusted post-hoc tests $(F(2,426)=58.52$, $p < 0.001$, $\eta_{G}^2=0.097)$
shows that, on average, significantly more use cases are correctly classified for zoomorphic and abstract gestures ($M=2.9$ for both modalities) than for light displays $(M=1.9)$.\\
In regards to the subjective evaluation, zoomorphic gestures are rated better (closer to the scale maximum) across the considered dimensions than the alternatives (see Fig. \ref{fig:ANOVA2}). Zoomorphic gestures are significantly more attractive $(F(2,426)=35.85$, $p < 0.001$, $\eta_{G}^2=0.049)$, 
provide significantly more joy when using 
$(F(2,426)=31.84$, $p < 0.001$, $\eta_{G}^2=0.034)$ 
and are significantly more intuitive to use 
$(F(2,426)=50.17$, $p < 0.001$, $\eta_{G}^2=0.072)$.
The intent to use is significantly higher for both gesture-based modalities than for the light display 
$(F(2,426)=27.55$, $p < 0.001$, $\eta_{G}^2=0.025)$.

Asked explicitly, which modality participants prefer for RtH communication in the given scenario, 56.1\% of the participants chose zoomorphic gestures for reasons such as "it is the most logical," "it is the most human," "appears likable". 21.1\% preferred abstract gestures ("easiest to understand," "clear and simple, zoomorphic gestures are too much," "modern"), while the remaining 16.8\% preferred light displays ("personal preference," "colors are more logical," "easy to understand once you know the light codes"). 

\newsec{Discussion}
Our results indicate that our zoomorphic and abstract gestures are more intuitively understood than our light displays.
Other work investigating zoomorphic gestures generally report higher accuracies (up to 75\%, e.g., \cite{Gacsi.2016}) than our results.
However, these studies investigate the communication of affective states, not functional states. 
As zoomorphic gestures are emblems, they are naturally closer related to affective information \cite{Knapp.1972}; hence, achieving high classification accuracy may be easier.
When considering the use cases individually, space for improvement exists, especially for the use cases "greet user" and "wait for new command" (see Fig. \ref{fig:Histo2}). Nonetheless, all robot states are identified above the chance level for each modality.

The subjective evaluation regarding user experience and intent to use is favorable toward zoomorphic gestures. Zoomorphic gestures are significantly more attractive and intuitive and provide more joy when using. 
\CHANGED{Compared to zoomorphic gestures, the abstract gestures achieved similarly high accuracies. Given that the participants had no prior training on any of the modalities, this suggests that our abstract gestures may not entail a higher learning effort than the other modalities, contrary to our expectations. 
Yet, the majority of participants stated a preference for zoomorphic gestures ($56.1\%$) over abstract ones ($21.2\%$).
The clear user preference for movement-based communication mirrors findings from previous studies \cite{Venture2019robot}.}

Overall, the sample used for this study included participants with different levels of experience with robots. However, the majority had limited experience with industrial cobots as very experienced users are hard to find. 
Therefore, our findings indicate that zoomorphic gestures may be both an intuitive and attractive modality, especially for inexperienced users. Similar to results by \cite{Aubert.2018}, comments made by some participants indicate that zoomorphic gestures may be perceived as annoying by more experienced users. The reason for this may be that they are more time-consuming and elaborate than the alternative modalities. The longer execution times of the gestures may impede the work-flow of assembly and, thus, may be less suitable for work processes with strict cycle times. 

\section{Limitations}
Our approach and methods also face some limitations.
To our knowledge, this work is the first to investigate zoomorphic gestures for appearance-constrained cobots. Therefore, we have followed an exploratory approach and consciously omitted safety aspects from consideration. 
However, especially for movement-based communication, safety is critical and requires extensive consideration.
These concerns need to be addressed in future studies.

Concerning our evaluation methods, our exploratory approach has some drawbacks. We worked with opportunity samples to recruit a large and diverse sample for both user studies to increase the generalizability of the results.
The same studies conducted only with highly experienced users may yield different results.

Further, the videos we produced of the different design options included less contextual information for the light displays than in the videos for the gesture-based modalities. In future studies, we advise maintaining an equal level of contextual information to reduce the possible influence of confounders. \CHANGED{Additionally, we suggest controlling for possible covariates, such as the total duration of the gestures.}
We also did not fully exhaust the design space of light displays in general. 
Instead, we built design options from \CHANGED{existing} guidelines under the consideration of the built-in hardware of the utilized cobot. 
\CHANGED{On a similar note, our design guidelines to develop different gesture and light options offer some leeway. Therefore, we designed study I to reduce arbitrariness within our design choices. Nonetheless, further studies with different design options and use cases are required to support the generalizability of our results.}

Finally, we recommend revising the benchmark study questionnaire
to include the participant's certainty of their classification. This additional information on certainty may provide helpful insights.

\section{Conclusions}
This paper's main objective is to 
explore the suitability of zoomorphic gestures for appearance-constrained cobots in industry applications. The two user studies indicate that the proposed guidelines for developing zoomorphic and abstract gestures are suitable. Participants understood both gesture-based modalities intuitively with an average classification accuracy of 58\%. Additionally, the subjective evaluation and stated preference are in favor of zoomorphic gestures.

Future research avenues include 
gesture development for additional use cases, the consideration of safety as outlined in \cite{ISO}, and  evaluation with experienced users. 
Further, zoomorphic gestures, which mimic the social dynamic between dogs and humans, may present 
\CHANGED{a compelling option} to reduce robot abuse by providing relevant social mechanisms \cite{Bartneck.Hu2008, Lucas.2016, Tan.2018}.
In a deployed system, zoomorphic gestures could also be combined with other visual and non-visual modalities, which were not considered in this paper as they represent orthogonal research directions. When considering unimodal communication, zoomorphic gestures may be especially suitable for novice users by lowering inhibitions. Further, zoomorphic gestures may also be made available in libraries provided in robot programming platforms 
to personalize the human-robot interaction.

\section*{Acknowledgements}
We would like to thank Elie Aljalbout and Konstantin Ritt for their technical support and Stefan Groß for supporting the preparation of the video materials.
We would also like thank r/aww for providing insights on dog body language.

\bibliographystyle{IEEEtran}
\bibliography{IEEEabrv, bibliography}

\begin{thebibliography}{10}
\providecommand{\url}[1]{#1}
\csname url@rmstyle\endcsname
\providecommand{\newblock}{\relax}
\providecommand{\bibinfo}[2]{#2}
\providecommand\BIBentrySTDinterwordspacing{\spaceskip=0pt\relax}
\providecommand\BIBentryALTinterwordstretchfactor{4}
\providecommand\BIBentryALTinterwordspacing{\spaceskip=\fontdimen2\font plus
\BIBentryALTinterwordstretchfactor\fontdimen3\font minus
  \fontdimen4\font\relax}
\providecommand\BIBforeignlanguage[2]{{%
\expandafter\ifx\csname l@#1\endcsname\relax
\typeout{** WARNING: IEEEtran.bst: No hyphenation pattern has been}%
\typeout{** loaded for the language `#1'. Using the pattern for}%
\typeout{** the default language instead.}%
\else
\language=\csname l@#1\endcsname
\fi
#2}}

\bibitem{Barattini.2012}
P.~Barattini, C.~Morand, and N.~M. Robertson, ``A proposed gesture set for the
  control of industrial collaborative robots,'' in \emph{RO-MAN}, 2012.

\bibitem{fulton2019robot}
M.~Fulton, C.~Edge, and J.~Sattar, ``Robot communication via motion: Closing
  the underwater human-robot interaction loop,'' in \emph{ICRA}, 2019.

\bibitem{BarakaVeloso.2018}
K.~Baraka and M.~M. Veloso, ``Mobile service robot state revealing through
  expressive lights. formalism, design, and evaluation,'' \emph{International
  Journal of Social Robotics}, 2018.

\bibitem{Aubert.2018}
M.~C. Aubert, H.~Bader, and K.~Hauser, ``Designing multimodal intent
  communication strategies for conflict avoidance in industrial human-robot
  teams,'' in \emph{RO-MAN}, 2018.

\bibitem{Onnasch.2016}
L.~Onnasch, X.~Maier, and T.~Jürgensohn, ``Mensch-roboter-interaktion - eine
  taxonomie für alle anwendungsfälle,'' \emph{baua: Fokus, Bundesanstalt für
  Arbeitsschutz und Arbeitsmedizin}, 2016.

\bibitem{Bethel.2007}
C.~L. Bethel and R.~R. Murphy, ``Survey of non-facial/non-verbal affective
  expressions for appearance-constrained robots,'' \emph{IEEE Transactions on
  Systems, Man, and Cybernetics, Part C (Applications and Reviews)}, 2007.

\bibitem{Syrdal.2010}
D.~S. Syrdal, K.~L. Koay, M.~G{\'a}csi, M.~L. Walters, and K.~Dautenhahn,
  ``Video prototyping of dog-inspired non-verbal affective communication for an
  appearance constrained robot,'' in \emph{RO-MAN}, 2010.

\bibitem{Kaetsyri.2015}
J.~K{\"a}tsyri, K.~F{\"o}rger, M.~M{\"a}k{\"a}r{\"a}inen, and T.~Takala, ``A
  review of empirical evidence on different uncanny valley hypotheses: support
  for perceptual mismatch as one road to the valley of eeriness,''
  \emph{Frontiers in Psychology}, 2015.

\bibitem{deshmukh2018more}
A.~Deshmukh, B.~Craenen, M.~E. Foster, and A.~Vinciarelli, ``The more i
  understand it, the less i like it: The relationship between understandability
  and godspeed scores for robotic gestures,'' in \emph{RO-MAN}, 2018.

\bibitem{Harrison.2012}
C.~Harrison, J.~Horstman, G.~Hsieh, and S.~Hudson, ``Unlocking the expressivity
  of point lights,'' in \emph{CHI}, 2012.

\bibitem{Cha.2017}
E.~Cha, T.~Trehon, L.~Wathieu, C.~Wagner, A.~Shukla, and M.~J. Matari{\'c},
  ``Modlight: designing a modular light signaling tool for human-robot
  interaction,'' in \emph{ICRA}, 2017.

\bibitem{Chadalavada.2015}
R.~T. Chadalavada, H.~Andreasson, R.~Krug, and A.~J. Lilienthal, ``That's on my
  mind! robot to human intention communication through on-board projection on
  shared floor space,'' in \emph{ECMR}, 2015.

\bibitem{Unhelkar.2014}
V.~V. Unhelkar, H.~C. Siu, and J.~A. Shah, ``Comparative performance of human
  and mobile robotic assistants in collaborative fetch-and-deliver tasks,'' in
  \emph{HRI}, 2014.

\bibitem{Baraka.2016}
K.~Baraka, A.~Paiva, and M.~Veloso, ``Expressive lights for revealing mobile
  service robot state,'' in \emph{Robot 2015: Second Iberian Robotics
  Conference}, 2016.

\bibitem{Szafir.2015}
D.~Szafir, B.~Mutlu, and T.~Fong, ``Communicating directionality in flying
  robots,'' in \emph{HRI}, 2015.

\bibitem{ElMakrini.2017}
I.~{El Makrini}, K.~Merckaert, D.~Lefeber, and B.~Vanderborght, ``Design of a
  collaborative architecture for human-robot assembly tasks,'' in \emph{IROS},
  2017.

\bibitem{DePace.2018}
F.~De~Pace, F.~Manuri, A.~Sanna, and D.~Zappia, ``An augmented interface to
  display industrial robot faults,'' in \emph{AVR}, 2018.

\bibitem{gleeson2013gestures}
B.~Gleeson, K.~MacLean, A.~Haddadi, E.~Croft, and J.~Alcazar, ``Gestures for
  industry intuitive human-robot communication from human observation,'' in
  \emph{HRI}, 2013.

\bibitem{Zanchettin.2013}
A.~M. Zanchettin, L.~Bascetta, and P.~Rocco, ``Acceptability of robotic
  manipulators in shared working environments through human-like redundancy
  resolution,'' \emph{Applied Ergonomics}, 2013.

\bibitem{Gasparetto2012Trajectory}
A.~Gasparetto, P.~Boscariol, A.~Lanzutti, and R.~Vidoni, ``Trajectory planning
  in robotics,'' \emph{Mathematics in Computer Science}, 2012.

\bibitem{Dragan.2015}
A.~D. Dragan, S.~Bauman, J.~Forlizzi, and S.~S. Srinivasa, ``Effects of robot
  motion on human-robot collaboration,'' in \emph{HRI}, 2015.

\bibitem{knight2016expressivediss}
H.~Knight, ``Expressive motion for low degree-of-freedom robots,'' 2016.

\bibitem{takahashi2010remarks}
K.~Takahashi, M.~Hosokawa, and M.~Hashimoto, ``Remarks on designing of
  emotional movement for simple communication robot,'' in \emph{2010 IEEE
  International Conference on Industrial Technology}, 2010.

\bibitem{fernandez2018passive}
R.~Fernandez, N.~John, S.~Kirmani, J.~Hart, J.~Sinapov, and P.~Stone, ``Passive
  demonstrations of light-based robot signals for improved human
  interpretability,'' in \emph{RO-MAN}, 2018.

\bibitem{Venture2019robot}
G.~Venture and D.~Kuli{\'c}, ``Robot expressive motions: a survey of generation
  and evaluation methods,'' \emph{ACM Transactions on Human-Robot Interaction
  (THRI)}, 2019.

\bibitem{Singh.2013}
A.~Singh and J.~E. Young, ``A dog tail for utility robots: exploring affective
  properties of tail movement,'' in \emph{IFIP Conference on Human-Computer
  Interaction}, 2013, pp. 403--419.

\bibitem{Gacsi.2016}
M.~G{\'a}csi, A.~Kis, T.~Farag{\'o}, M.~Janiak, R.~Muszy{\'n}ski, and
  {\'A}.~Mikl{\'o}si, ``Humans attribute emotions to a robot that shows simple
  behavioural patterns borrowed from dog behaviour,'' \emph{Computers in Human
  Behavior}, 2016.

\bibitem{Young.2011}
J.~E. Young, Y.~Kamiyama, J.~Reichenbach, T.~Igarashi, and E.~Sharlin, ``How to
  walk a robot: A dog-leash human-robot interface,'' in \emph{RO-MAN}, 2011.

\bibitem{Knapp.1972}
M.~L. Knapp, \emph{Nonverbal communication in human interaction}.\hskip 1em
  plus 0.5em minus 0.4em\relax {Holt Rinehart and Winston}, 1972.

\bibitem{akgun2012trajectories}
B.~Akgun, M.~Cakmak, J.~W. Yoo, and A.~L. Thomaz, ``Trajectories and keyframes
  for kinesthetic teaching: A human-robot interaction perspective,'' in
  \emph{HRI}, 2012.

\bibitem{bartneck2009measurement}
C.~Bartneck, D.~Kuli{\'c}, E.~Croft, and S.~Zoghbi, ``Measurement instruments
  for the anthropomorphism, animacy, likeability, perceived intelligence, and
  perceived safety of robots,'' \emph{International Journal of Social
  Robotics}, 2009.

\bibitem{Weiss.2015}
A.~Weiss and C.~Bartneck, ``Meta analysis of the usage of the godspeed
  questionnaire series,'' in \emph{RO-MAN}, 2015.

\bibitem{Fanning.2005}
E.~Fanning, ``Formatting a paper-based survey questionnaire: Best practices,''
  \emph{Practical Assessment, Research, and Evaluation}, 2005.

\bibitem{Schrepp}
M.~Schrepp and J.~Thomaschewski, ``Handbook for the modular extension of the
  user experience questionnaire,'' in \emph{Mensch \& Computer}, 2019.

\bibitem{Harborth.2018}
D.~Harborth and S.~Pape, ``German translation of the unified theory of
  acceptance and use of technology 2 {(UTAUT2)} questionnaire,'' \emph{SSRN
  Journal}, 2018.

\bibitem{ISO}
ISO, ``{TS} 15066: 2016: Robots and robotic devices -- collaborative robots,''
  \emph{International Organization for Standardization}, 2016.

\bibitem{Bartneck.Hu2008}
C.~Bartneck and J.~Hu, ``Exploring the abuse of robots,'' \emph{Interaction
  Studies}, 2008.

\bibitem{Lucas.2016}
H.~Lucas, J.~Poston, N.~Yocum, Z.~Carlson, and D.~Feil-Seifer, ``Too big to be
  mistreated? examining the role of robot size on perceptions of
  mistreatment,'' in \emph{RO-MAN}, 2016.

\bibitem{Tan.2018}
X.~Z. Tan, M.~V{\'a}zquez, E.~J. Carter, C.~G. Morales, and A.~Steinfeld,
  ``Inducing bystander interventions during robot abuse with social
  mechanisms,'' in \emph{HRI}, 2018.

\end{thebibliography}

\end{document}